\documentclass[a4paper]{cas-dc}
\usepackage{cite}
\usepackage[numbers]{natbib}
\usepackage{hhline}
\usepackage{threeparttable}
\usepackage{ulem}

\begin{document}
\title[mode=title]{E3CM: Epipolar-Constrained Cascade Correspondence Matching}                      
\shorttitle{E3CM: Epipolar-Constrained Cascade Correspondence Matching}
\shortauthors{Chenbo Zhou et~al.}

\tnotetext[1]{This work was supported by the National Key R\&D Program of China under Grant 2020AAA0108100, the National Natural Science Foundation
of China under Grant 62233013, the Science and Technology Commission of Shanghai Municipal under Grant 22511104500, and the Fundamental
Research Funds for the Central Universities.} 

\author{Chenbo Zhou}[orcid=0000-0002-0736-6086]
\author{Shuai Su}[orcid=0000-0001-6144-8923]
\author{Qijun Chen}[orcid=0000-0001-5644-1188]
\author{Rui Fan}[orcid=0000-0003-2593-6596]
\cormark[1]
\affiliation{Tongji University,
city={Shanghai}, postcode={201804}, country={China}}
\cortext[cor1]{Corresponding author, email address: rui.fan@ieee.org}

\begin{abstract}
Accurate and robust correspondence matching is of utmost importance for various 3D computer vision tasks. However, traditional explicit programming-based methods often struggle to handle challenging scenarios, and deep learning-based methods require large well-labeled datasets for network training. In this article, we introduce Epipolar-Constrained Cascade Correspondence (E3CM), a novel approach that addresses these limitations. Unlike traditional methods, E3CM leverages pre-trained convolutional neural networks to match correspondence, without requiring annotated data for any network training or fine-tuning. Our method utilizes epipolar constraints to guide the matching process and incorporates a cascade structure for progressive refinement of matches. We extensively evaluate the performance of E3CM through comprehensive experiments and demonstrate its superiority over existing methods. To promote further research and facilitate reproducibility, we make our source code publicly available at \url{https://mias.group/E3CM/}.
\end{abstract}


\maketitle

\begin{figure*}
\centering
\includegraphics[width=0.99\textwidth]{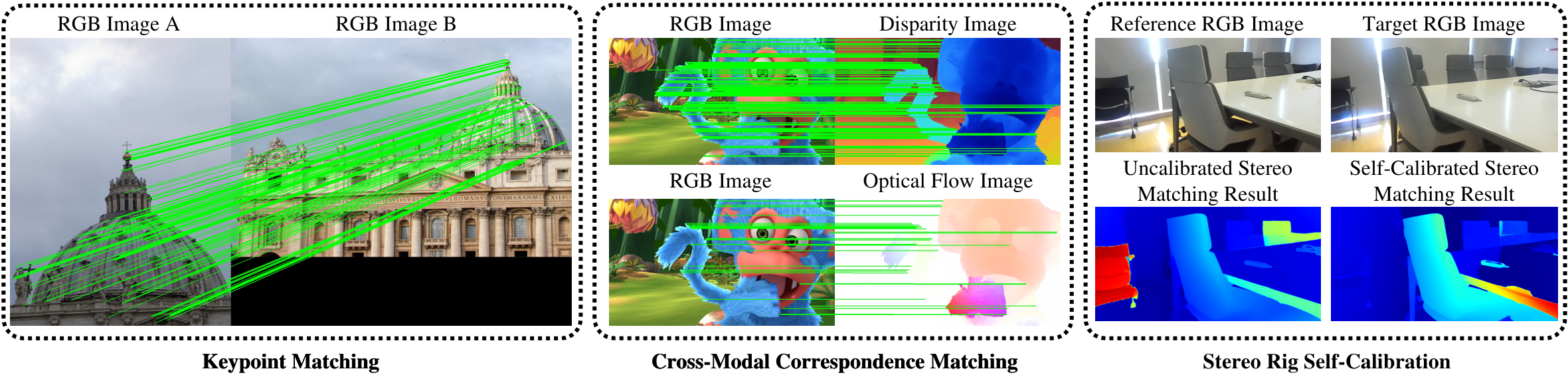}
\caption{Our proposed E3CM algorithm can be applied to various 3D computer vision applications, including: (a) keypoint matching, (b) cross-modal correspondence matching, and (c) online stereo rig self-calibration.}
\label{fig:firstpage}		
\end{figure*}

\section{Introduction}
\label{sec.intro}

Correspondence matching forms the foundation for a variety of 3D computer vision tasks, such as simultaneous localization and mapping (SLAM) \cite{engel2017direct,mur2015ORB,yu2022accurate}, structure from motion (SfM) \cite{schonberger2016structure, wu2013towards, fan2023autonomous}, dense disparity estimation and transformation \cite{fan2019pothole, fan2021graph, fan2019road}, and online stereo rig self-calibration \cite{ling2016high, wu2021simultaneous}. Currently, popular correspondence matching algorithms are classified into two categories: i) explicit programming-based and ii) deep learning-based.

Explicit programming-based correspondence matching methods typically extract keypoints based on human-defined local features, such as rotated binary robust independent elementary features (ORB) \cite{rublee2011ORB} and scale-invariant feature transform (SIFT) \cite{lowe2004distinctive}, followed by making pairs between correspondences with nearest neighboring (NN) matching \cite{muja2014scalable}. In contrast, recent deep learning-based methods \cite{detone2018SuperPoint,dusmanu2019d2,revaud2019r2d2} learn to detect and describe local features with neural networks, resulting in significantly enhanced robustness in correspondence matching when compared to explicit programming-based techniques. Furthermore, matchers based on graph neural networks \cite{sarlin2020SuperGlue} have been developed and utilized to predict correct matches while effectively filtering out incorrect ones.

In scenarios involving large perspective changes or repetitive textures, explicit programming-based methods tend to exhibit subpar performance. However, deep learning-based methods, as demonstrated in \cite{detone2018SuperPoint,dusmanu2019d2,revaud2019r2d2}, have made remarkable strides in improving matching accuracy within such challenging scenarios. Nevertheless, it is worth noting that the existing deep learning-based methods often rely on a large amount of labeled training data, which typically includes precise information regarding camera positions and poses. To address the limitations associated with these methods, a novel approach, referred to as deep feature matching (DFM), was introduced \cite{efe2021dfm}. DFM employs a hierarchical matching strategy based on the Visual Geometry Group (VGG) network, pre-trained on the ImageNet \cite{krizhevsky2012imagenet} database. An intriguing aspect of DFM is that it does not necessitate additional training using data annotated with correspondences, thus mitigating the need for a large amount of labeled data. Furthermore, DFM offers significantly improved accuracy and robustness compared to explicit programming-based methods, surpassing even some approaches trained with correspondences. However, it is important to note that DFM's hierarchical correspondence matching process is based on homography matrices \cite{fan2021learning}, which imposes a limitation on its applicability. Specifically, DFM is most effective when dealing with cases that involve only planar surfaces. In scenarios where the scene contains non-planar or complex surfaces, the performance of DFM may be compromised. The reliance on homography matrices can lead to suboptimal performance in more intricate stereoscopic scenes, especially indoor scenes with short sight distances. 

Hence in this paper, we introduce \uline{\textbf{E}pipolar-\textbf{C}onstrained \textbf{C}ascade \textbf{C}orrespondence \textbf{M}atching (\textbf{E3CM})}, a plug-and-play solution designed specifically for stereo rigs. Our method leverages feature maps derived from pre-trained backbones (on the ImageNet database), incorporating a cascade outlier rejection module that relies on pose estimation and epipolar constraints. This combination of techniques enables E3CM to effectively handle the typical and prevalent scenarios encountered in real-world stereoscopic scenes. The matching process begins from the final layer of multiple selected feature maps. We utilize the obtained matches to estimate the camera's pose. 
Using the estimated pose, we then apply the epipolar constraint to eliminate outliers in the previous layer. Subsequently, a new pose is estimated based on the matches after outlier removal. This iterative process continues until reaching the first layer, gradually enhancing the pose estimation accuracy and increasing the number of correct matches. Additionally, we extend the utilization of pre-trained backbones, enabling a comprehensive comparison among various backbones. To further enhance the reliability of matches within the feature maps, we introduce a novel confidence score, which effectively decreases the probability of incorrectly estimating camera poses, further enhancing the overall performance of E3CM. As depicted in Fig. 1, E3CM can be effectively utilized for cross-modal correspondence matching as well as various 3D computer vision tasks, exemplified with online stereo rig self-calibration in this paper.

The structure of the remaining paper is as follows: Section \ref{sec.related_work} provides an overview of existing works related to correspondence matching, outlining the advancements and limitations in the field. Section \ref{sec.method} presents the details of our proposed E3CM algorithm, explaining the key components and techniques employed. The experimental results for performance evaluation are illustrated
in Section \ref{sec.experiments}. Section \ref{sec.discussion} delves into a comprehensive discussion of the applicability of our work, addressing its potential applications and limitations. Finally, Section \ref{sec.conclusion} serves as a summary of the paper, highlighting the key findings and contributions.

\section{Related Work}
\label{sec.related_work}
Correspondence matching is a crucial task in 3D computer vision applications, such as visual odometry, image stitching, and online stereo rig self-calibration. Traditional explicit programming-based correspondence matching approaches \cite{rublee2011ORB,lowe1999object,bay2006surf,alcantarilla2011fast} are typically based on hand-crafted techniques. These approaches primarily rely on local gradients, local corners, and local blob features to detect and describe keypoints. However, they often struggle to perform well in low-texture scenes, which makes them less reliable for downstream applications.
	
Recent advancements in deep learning-based approaches for keypoint detection, description, and matching have demonstrated superior accuracy and robustness compared to traditional methods. SuperPoint \cite{detone2018SuperPoint} is a notable method in the field of keypoint detection and description. It is a self-supervised detector-descriptor framework that initially introduces a detector referred to as MagicPoint using synthetic images and then trains a descriptor by generating random homography matrices as ground truth. This training strategy has been widely adopted in many other works. Another deep learning-based method, D2-Net \cite{dusmanu2019d2}, is a trainable convolutional neural network (CNN) that performs joint detection and description of local features. It extracts keypoints by computing the maximum values on a feature map that is four times smaller than the source image. However, D2-Net prioritizes repeatability, which leads to reduced matching performance in regions with high texture repetition. In response to this limitation, R2D2 \cite{revaud2019r2d2} builds upon D2-Net by emphasizing the reliability of feature points.
	
While NN matching has been widely utilized in the matching stage, it often overlooks the assignment structure and disregards visual information. SuperGlue \cite{sarlin2020SuperGlue} presents a novel approach to tackle this issue. It leverages a graph neural network (GNN) with an attention mechanism to integrate positional information and keypoint descriptions, and computes the correspondences using the Sinkhorn algorithm \cite{cuturi2013sinkhorn,sinkhorn1967concerning}.

\begin{figure*}
	\centering
	\includegraphics[width=0.99\textwidth]{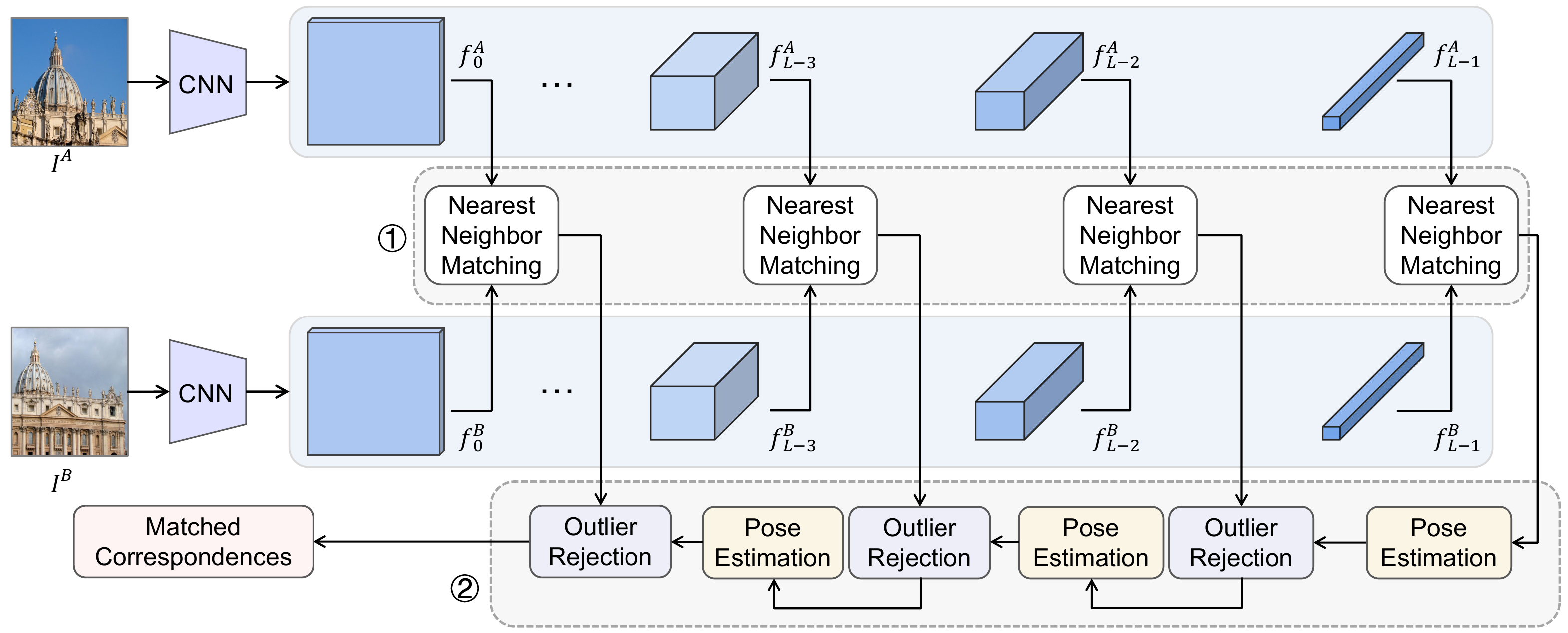}
	\caption{The framework of our proposed epipolar-constrained cascade correspondence matching approach: \textcircled{1} initial deep correspondence matching; \textcircled{2} epipolar-constrained cascade refinement. }
	\label{fig:wholestructure}
\end{figure*}
	
In addition to the traditional two-stage pipeline consisting of both a detector and descriptor, several existing works have adopted end-to-end frameworks for correspondence matching. Neighborhood consensus networks (NCNet) \cite{rocco2018neighbourhood} were proposed to match dense correspondences without the need for a separate feature detector. However, due to intensive correlation score computation on down-scaled feature maps, NCNet's performance in camera pose estimation tasks is suboptimal. To address this limitation, efficient neighborhood consensus network via submanifold sparse convolutions (SparseNCNet) \cite{rocco2020efficient} employs a sparse representation of the correlation tensor by storing a portion of the scores and replacing dense 4D convolutions with sparse convolutions. Densely connected recurrent convolutional neural network (DRC-Net) \cite{li2020dual} follows SparseNCNet and introduces a hierarchical framework to generate dense matches with higher accuracy. An epipolar-guided pixel-level correspondence matching approach, referred to as Patch2pix \cite{zhou2021patch2pix}, leverages pre-trained backbones to extract potential patch-level matches and refines the matches to pixel-level accuracy using two-stage regressors. Another detector-free framework, local feature matching with Transformers (LoFTR) \cite{sun2021loftr}, builds upon the Transformer network and achieves superior performance compared to SuperPoint with SuperGlue.

Furthermore, various strategies have been proposed to improve the accuracy of matches by effectively rejecting outliers. Neural-guided random sample consensus (RANSAC) \cite{brachmann2019neural} utilizes probabilities to weigh the matches. AdaLAM \cite{cavalli2020adalam} assumes that close matching pairs share the same local affine transformation and rejects outliers that deviate from the affine matrix within a neighboring area. Another approach, presented in \cite{yi2018learning}, employs neural networks to predict binary labels for outlier identification.

The work presented in \cite{efe2021dfm} can be considered as a baseline for the task of matching keypoints using a pre-trained backbone network. It employs a coarse-to-fine strategy to match features from deep layers to shallow layers. Additionally, it also estimates homography matrices using feature maps to refine correspondences. However, their approach is limited to cases that involve only planar surfaces. In our work, we address this limitation by introducing the epipolar-constrained cascade refinement strategy, which replaces the two-stage method and enables matching in more general scenarios. Additionally, this paper evaluates the effectiveness of our proposed E3CM approaches using various popular backbone networks, including VGG \cite{simonyan2014very}, ResNet \cite{he2016deep}, DenseNet \cite{huang2017densely}, MobileNet \cite{howard2017mobilenets}, and GoogleNet \cite{szegedy2015going}. These backbone networks have all been pre-trained on the ImageNet database.

\section{Methodology}
\label{sec.method}

\subsection{Initial Deep Correspondence Matching}
	\label{sec.methodology_deep_correspondence_matching}
	
Our proposed initial deep correspondence matching approach is developed based on DFM \cite{efe2021dfm}, where a VGG-19 \cite{simonyan2014very} network pre-trained on the ImageNet database is employed to extract deep feature maps, $\boldsymbol{f}^{A}$ and $\boldsymbol{f}^{B}$, from a given pair of color images, ${I}^A$ and ${I}^B$. Based on the hypothesis that the given image pairs can be linked by a homography matrix $\boldsymbol{H}^{BA}$, DFM warps the color images using an estimated homography matrix before the second-stage correspondence matching. However, this limits its applicability to scenarios that primarily involve a planar surface. Additionally, DFM only demonstrates its compatibility with pre-trained VGG models, as it requires the resolution of the first feature map to be identical to that of the input images. To address these limitations, we extend the applicability of the method to general cases by removing the redundant image warping step. Moreover, we enhance its compatibility with other state-of-the-art CNNs to broaden its usage and accommodate different network architectures.
	
Correspondence matching at layer $l$ can be solved with NN matching. Let $\boldsymbol{f}_{l}^{A}$ and $\boldsymbol{f}_{l}^{B}$ be the feature maps (size: $ H / 2^{l} \times W / 2^{l} \times C_{l}$ ) at layer $l\in[0, L-1]$ extracted from a given pair of images ${I}^A$ and ${I}^B$ (resolution: $H \times W$ pixels). Given a pair of points $\boldsymbol{p}_l^{A}=(h_l^{A};w_l^{A})$ in $\boldsymbol{f}_{l}^{A}$ and $\boldsymbol{p}_l^{B}=(h_l^{B};w_l^{B})$ in $\boldsymbol{f}_{l}^{B}$, where $0<h_l^{A,B}<H / 2^{l}$ and $0<w_l^{A,B}<W / 2^{l}$, we measure the distance $d$ between their representations $\boldsymbol{c}(\boldsymbol{p}_l^{A})$ and $\boldsymbol{c}(\boldsymbol{p}_l^{B})$ of size: $C_{l}\times 1$ as follows:
\begin{equation}
d(\boldsymbol{p}_l^{A},\boldsymbol{p}_l^{B})=1-\Phi(\boldsymbol{c}({\boldsymbol{p}_l^{A}),\boldsymbol{c}(\boldsymbol{p}_l^{B}})),
\label{eq.d}
\end{equation}
where
\begin{equation}
\Phi(\boldsymbol{c}({\boldsymbol{p}_l^{A}),\boldsymbol{c}(\boldsymbol{p}_l^{B}}))=\frac{\boldsymbol{c}({\boldsymbol{p}_l^{A})}^{\top} \boldsymbol{c}({\boldsymbol{p}_l^{B})}}{\big|\big|\boldsymbol{c}({\boldsymbol{p}_l^{A}})\big|\big|_2 \big|\big|\boldsymbol{c}({\boldsymbol{p}_l^{B}})\big|\big|_2}.
\end{equation}is the cosine distance between $\boldsymbol{c}(\boldsymbol{p}_l^{A})$ and $\boldsymbol{c}(\boldsymbol{p}_l^{B})$. A map $\boldsymbol{D}_l$ (size: $ H / 2^{l} \times W / 2^{l} \times ( H / 2^{l} \times W / 2^{l})$) storing the measured cosine distances between all possible matches can thus be obtained. Given $\boldsymbol{p}_{{l}_{k}}^{A}$, if the ratio of its minimum distance (corresponding to $\boldsymbol{p}_{{l}_{i}}^{B}$) versus its second-minimum distance (corresponding to $\boldsymbol{p}_{{l}_{j}}^{B}$) is lower than the pre-set threshold (empirically set to 0.9), $(\boldsymbol{p}_{{l}_{k}}^{A},\boldsymbol{p}_{{l}_{i}}^{B})$ are considered as a pair of satisfactorily matched correspondences. 

Compared to the traditional approaches that determine correspondences by detecting, describing, and matching local visual features via explicit programming, our proposed CNN-based approach, on the other hand, can directly perform correspondence matching on deep hierarchical feature maps. One of the most representative characteristics of CNNs is that the feature maps at shallow layers have higher resolutions and smaller receptive fields, while the feature maps at deeper layers have lower resolutions and larger receptive fields. Therefore, more confident but fewer correspondences can be obtained when it comes to the deeper layers of the CNNs. Based on this important characteristic, we develop an epipolar-constrained cascade refinement strategy for outlier rejection, as presented in the following subsection. 
	
		\begin{figure}[t!]
		\centering
		\includegraphics[width=0.499\textwidth]{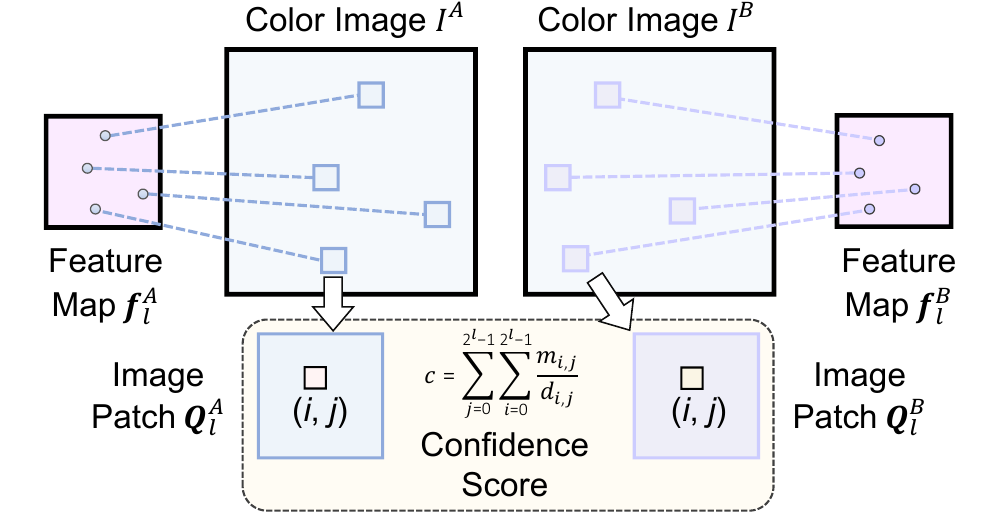}
		\caption{Confidence score computation. 
%			The confidence scores of the matching pairs in the feature maps are related to the matching situation in the corresponding patches in the source image.
			}
		\label{fig:confidencescore}
	\end{figure}

	\subsection{Epipolar-Constrained Cascade Refinement}
	
	As illustrated in Fig \ref{fig:confidencescore}, given a pair of matched points $\boldsymbol{p}_l^A$ and $\boldsymbol{p}_l^B$, respectively in the feature maps $\boldsymbol{f}_{l}^A$ and  $\boldsymbol{f}_{l}^B$, they correspond to two patches $\boldsymbol{Q}_l^A=\{\boldsymbol{q}^A_{i,j}\}_{i=0,j=0}^{2^{l}-1,2^{l}-1}$ and  $\boldsymbol{Q}_l^B=\{\boldsymbol{q}^B_{i,j}\}_{i=0,j=0}^{2^{l}-1,2^{l}-1}$ of size $2^{l} \times 2^{l}$ in the original images $I^A$ and $I^B$. 
	We define a confidence score
	\begin{equation}
        \centering
	c =\sum_{i=0}^{2^{l}-1}\sum_{j=0}^{2^{l}-1} \frac{m_{i, j}}{d_{i, j}}
	\label{eq.confidence_score}
	\end{equation}
	to measure the reliability of the matched image patches $\boldsymbol{Q}_l^A=\{\boldsymbol{q}^A_{i,j}\}_{i=0,j=0}^{2^{l}-1,2^{l}-1}$ and  $\boldsymbol{Q}_l^B=\{\boldsymbol{q}^B_{i,j}\}_{i=0,j=0}^{2^{l}-1,2^{l}-1}$, where $d_{i, j}$ represents the distance between $\boldsymbol{q}^A_{i,j}$ and $\boldsymbol{q}^B_{i,j}$, measured using the feature maps at the shallowest layer\footnote{An image might have to be resampled so that its resolution is identical to that of the feature map at the shallowest layer.}, $m_{i,j}=\{0,1\}$ is determined by NN matching (a good match corresponds to 1, while a bad match corresponds to 0). 
 
\begin{figure*}
		\centering
		\includegraphics[width=0.99\textwidth]{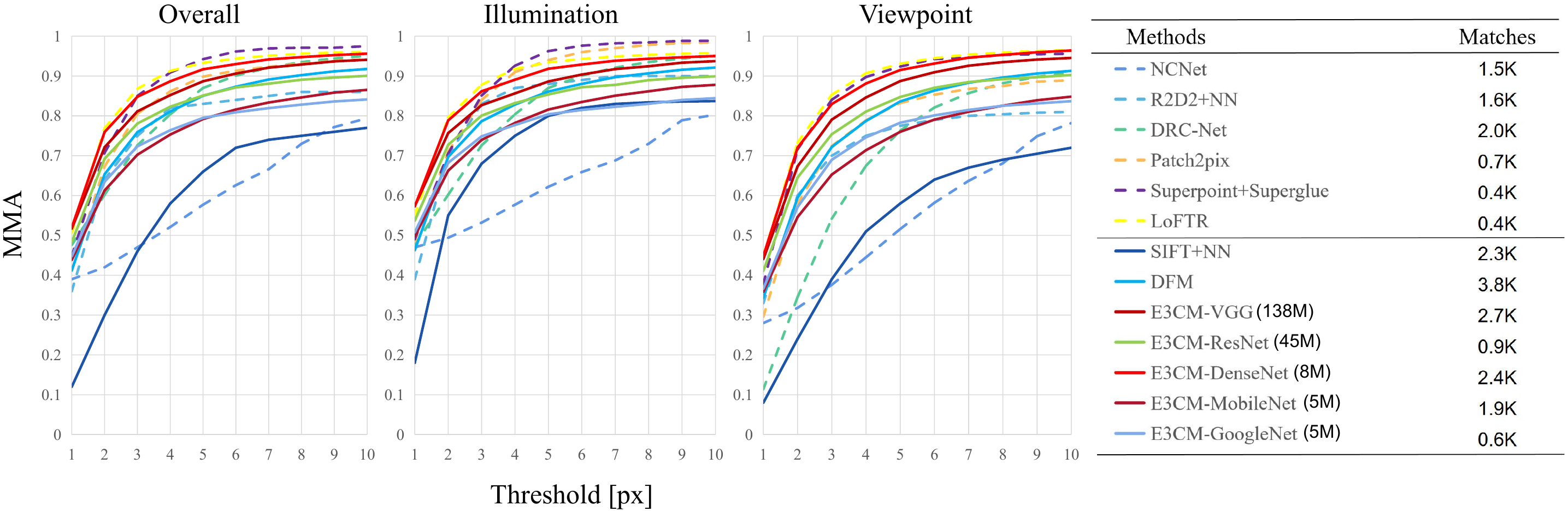}
		\caption{MMA comparison of different methods on the HPatches dataset. The methods above the dividing line in the table on the right are fully supervised ones, while the methods below the dividing line are training-free and plug-and-play ones.  }
		\label{fig:hpatchesmatching}
	\end{figure*}
 
Since the camera pose between $I^A$ and $I^B$ can be estimated using at least eight pairs of matched correspondences, we choose eight pairs of image patches with the highest confidence scores to compute the fundamental matrix $\boldsymbol{F}_l$ with respect to the $l$-th layer, where the centering point of each desired image patch is used in the eight-point algorithm \cite{hartley2003multiple}. The fundamental matrix $\boldsymbol{F}_l$ estimated using $\boldsymbol{f}_{l}^A$ and $\boldsymbol{f}_{l}^B$ is used to reject outliers (not satisfying the epipolar constraint) for correspondence matching using $\boldsymbol{f}_{l-1}^A$ and $\boldsymbol{f}_{l-1}^B$. In this paper, the Sampson distance \cite{hartley2003multiple}
	\begin{equation}
		\phi_{\boldsymbol{p}^{A}_l, \boldsymbol{p}_l^{B}}=\frac{\left(\left(\boldsymbol{p}_l^{B}\right)^{\top} \boldsymbol{F}_l \boldsymbol{p}_l^{A}\right)^{2}}{\left(\boldsymbol{F}_l \boldsymbol{p}_l^{A}\right)_{1}^{2}+\left(\boldsymbol{F}_l \boldsymbol{p}_l^{A}\right)_{2}^{2}+\left(\boldsymbol{F}_l^{\top} \boldsymbol{p}_l^{B}\right)_{1}^{2}+\left(\boldsymbol{F}_l^{\top} \boldsymbol{p}_l^{B}\right)_{2}^{2}}
	\end{equation}
	is used to determine inliers and outliers, 	where $\left(\boldsymbol{F} \boldsymbol{p}^{A}\right)_{k}^{2}$ and $\left(\boldsymbol{F} \boldsymbol{p}^{B}\right)_{k}^{2}$ represent the square of the $k$-th entry of the vector $\boldsymbol{F} \boldsymbol{p}^{A}$ and $\boldsymbol{F} \boldsymbol{p}^{B}$, respectively. As illustrated in Fig. \ref{fig:wholestructure}, such an outlier rejection algorithm performs iteratively from the deepest layer to the shallowest layer.

\section{Experiments}
\label{sec.experiments}
\subsection{Image matching} 
	
 	\begin{table*}
		\centering
		\fontsize{6}{13.5}\selectfont
				\begin{tabular}{cclcccccc}
					\hline
					\multirow{2}{*}{Category}           & \multicolumn{2}{c}{\multirow{2}{*}{Method}} & \multicolumn{6}{c}{Homography estimation accuracy}        \\ \cline{4-9} 
					& \multicolumn{2}{c}{}                        &  & \textless{}1px &  & \textless{}3px &  & \textless{}5px \\ 
					\toprule
                           \multirow{1}{*}{training-free}    & \multicolumn{2}{c}{SIFT \cite{lowe2004distinctive}+NN \cite{muja2014scalable}}               &  & 0.36           &  & 0.76           &  & 0.85  \\
                          \hline
					\multirow{9}{*}{fully supervised}   & \multicolumn{2}{c}{NCNet \cite{rocco2018neighbourhood}}           &  & 0.48           &  & 0.61           &  & 0.71          \\
					
					%\cline{2-9}
                    & \multicolumn{2}{c}{LIFT \cite{yi2016lift}}                 &  & 0.39           &  & 0.73           &  & 0.78       \\ 
                    
					& \multicolumn{2}{c}{R2D2 \cite{revaud2019r2d2}+NN \cite{muja2014scalable}}                 &  & 0.47           &  & 0.78           &  & 0.83           \\ 
					%\cline{2-9} 
					& \multicolumn{2}{c}{DRC-Net \cite{li2020dual}}                &  & 0.46           &  & 0.66           &  & 0.77           \\ 
                    & \multicolumn{2}{c}{SOSNet \cite{tian2019sosnet}}                 &  & 0.52           &  & 0.81           &  & 0.86     \\ 
                    & \multicolumn{2}{c}{MAGSAC \cite{barath2019magsac}}                 &  & 0.51           &  & 0.79           &  & 0.84     \\
					& \multicolumn{2}{c}{Pacth2pix \cite{zhou2021patch2pix}}           &  & 0.51           &  & 0.79           &  & 0.86           \\ 
					%\cline{2-9} 
					
					&\multicolumn{2}{c}{SuperPoint \cite{detone2018SuperPoint}+SuperGlue \cite{sarlin2020SuperGlue}}    &  & 0.53           &  & 0.84           &  & 0.90         \\
					& \multicolumn{2}{c}{LoFTR \cite{sun2021loftr}}           &  & $\textbf{0.66}$           &  & $\textbf{0.86}$           &  & $\textbf{0.92}$             \\
					\hline
					
                          \multirow{2}{*}{plug-and-play} 
   &  
					 \multicolumn{2}{c}{DFM \cite{efe2021dfm}}                     &  & 0.41           &  & 0.74           &  & 0.86           \\ 
					%\cline{2-9} 
					& \multicolumn{2}{c}{E3CM (Ours)}                    &  & \textcolor{blue}{\textbf{0.49}}          &  & \textcolor{blue}{\textbf{0.78}}           &  & \textcolor{blue}{\textbf{0.88}}           
					      \\ \hline 
			\end{tabular}
                \caption{\textbf{Accuracy of Homography Matrix Estimation on the HPatches dataset.} The table presents the percentages of correctly estimated homographies with average corner error distances below 1/3/5 pixels.}
		\medskip

		\label{table:homographyestimation}
	\end{table*}
	
	We first conduct experiments on the HPatches dataset \cite{balntas2017hpatches} to evaluate the performance of our correspondence matching method. The dataset consists of two main parts: `Viewpoint' and `Illumination'.
	
	As shown in Fig. \ref{fig:hpatchesmatching}, our method using DenseNet161 as the backbone outperforms SuperPoint+SuperGlue (trained with correspondences) under low threshold values in all three situations: `Overall', `Illumination', and `Viewpoint'. While our method may not achieve the best performance in the `Illumination' scenario, it excels in another two situations, particularly in the `Viewpoint' scenario. Additionally, our method produces a significantly higher number of matches compared to other methods, resulting in higher mean matching accuracy (MMA) scores.
	
	We also evaluate the performance of our proposed method with respect to different backbones, including VGG19, DenseNet161, ResNet152, MobileNet-Large, and GoogleNet. Among these backbones, DenseNet161 achieved the best performance. To ensure a fair comparison with DFM, we use VGG19 as the backbone for the following experiments.
	
	We analyze the network structures and speculated about possible reasons for the observed performance differences. ResNet, with its residual network structure, may hinder the flow of information between layers, leading to the loss of low-dimensional feature information, which is not ideal for using feature map channels directly as descriptors for feature matching. As for MobileNet-Large and GoogleNet, their lighter network structures inherently result in a loss of performance compared to more complex networks.

\subsection{Homography matrix estimation}
	
In the evaluation on the HPatches dataset, we estimate the homography matrix using our proposed E3CM method and compare it with the ground-truth homography matrix provided by the dataset. We select the four corners of the image and project them using both the ground-truth and estimated homography matrices. The average distance between the projected corner points is then calculated, and we evaluate the method using thresholds of (1, 3, 5). This allows us to determine the percentage of image pairs that fall below each threshold.

As shown in Table \ref{table:homographyestimation}, while our E3CM method may not perform as well as the state-of-the-art methods, it outperforms many methods that require training with correspondences. Furthermore, our method achieves the best performance in homography matrix estimation among the plug-and-play and training-free methods.

	\subsection{Pose estimation}
	\begin{table*}
		\centering
		\fontsize{6}{13.5}\selectfont
				\begin{tabular}{cclccccccll} 
					\hline
					\multirow{2}{*}{Category}           & \multicolumn{2}{c}{\multirow{2}{*}{Method}} & \multicolumn{6}{c}{Pose Estimation AUC}                                                                                                     &  & \multirow{2}{*}{P} \\ \cline{4-9} 
					& \multicolumn{2}{c}{}                        &                      & @5°                       &                      & @10°                      &                      & @20°                      &  &                       \\ \hline

                          \multirow{1}{*}{training-free} & \multicolumn{2}{c}{SIFT \cite{lowe2004distinctive}+NN \cite{muja2014scalable}}                    &                                         &4.78 &  & 10.71 &  & 20.44 &  & 17.19             \\       
                            \hline
					
					\multirow{9}{*}{fully supervised}  & \multicolumn{2}{c}{NCNet \cite{rocco2018neighbourhood}}                 &                      & 4.82                     &                      & 11.31                     &                      & 22.96                     &  & 54.85               \\
					& \multicolumn{2}{c}{LIFT \cite{yi2016lift}}                 &  & 6.03           &  & 13.71           &  & 27.96       &  & 39.97      \\ 
                   
					& \multicolumn{2}{c}{R2D2 \cite{revaud2019r2d2}+NN \cite{muja2014scalable}}                 &                      & 35.07                     &                      & 52.83                     &                      & 68.03                     &  & 81.33                 \\
					& \multicolumn{2}{c}{DRC-Net \cite{li2020dual}}                 &                      & 31.18                     &                      & 47.81                     &                      & 62.80                     &  & 85.72                 \\
                     & \multicolumn{2}{c}{SOSNet \cite{tian2019sosnet}}                 &  & 40.16           &  & 56.45           &  & 72.83       &  & 82.47      \\ 
                    & \multicolumn{2}{c}{MAGSAC \cite{barath2019magsac}}                 &  & 43.98           &  & 55.74           &  & 68.18       &  & 81.03      \\
					& \multicolumn{2}{c}{Patch2pix \cite{zhou2021patch2pix}}                 &                      & 43.32                     &                      & 58.34                     &                      & 70.27                     &  & 83.06                 \\
					
					&\multicolumn{2}{c}{SuperPoint \cite{detone2018SuperPoint}+SuperGlue \cite{sarlin2020SuperGlue}}    &                      & 42.28                     &                      & 62.36                     &                      & 77.86                     &  & 93.34                 
					\\
					& \multicolumn{2}{c}{LoFTR \cite{sun2021loftr}}                   &                      & $\textbf{52.80}$                     &                      & $\textbf{69.19}$                     &                      & $\textbf{81.18}$                     &  & $\textbf{97.18}$ \\  \hline

                           \multirow{2}{*}{plug-and-play} & \multicolumn{2}{c}{DFM \cite{efe2021dfm}}                     &                      & 35.17                     &                       & 50.64                     &                      & 61.12                     &  & 77.19                 \\
					& \multicolumn{2}{c}{E3CM(ours)}                    &                      & \textcolor{blue}{\textbf{39.85}}                    &                      & \textcolor{blue}{\textbf{54.11}}                    &                      & \textcolor{blue}{\textbf{65.86}}                     &  & \textcolor{blue}{\textbf{91.14}}              \\ \hline
			\end{tabular}
                \caption{\textbf{Evaluation on MegaDepth:} The table presents the percentages of correctly estimated poses with pose errors below 5/10/20 degrees. `P' refers to the matching precision.}
		\label{table:MegaDepth}
	\end{table*}

 	\begin{table*}
		\centering
		\fontsize{6}{13.5}\selectfont
				\begin{tabular}{cclccccccll} 
					\hline
					\multirow{2}{*}{Category}           & \multicolumn{2}{c}{\multirow{2}{*}{Method}} & \multicolumn{6}{c}{Pose Estimation AUC}                                                                                                     &  & \multirow{2}{*}{P} \\ \cline{4-9}
					& \multicolumn{2}{c}{}                        &                      & @5°                       &                      & @10°                      &                      & @20°                      &  &                       \\ \hline
                        \multirow{1}{*}{training-free} & \multicolumn{2}{c}{SIFT \cite{lowe2004distinctive}+NN \cite{muja2014scalable}}                    & &4.67 &  & 12.04 &  & 24.33 &  & 12.04    \\\hline
					\multirow{9}{*}{fully supervised}   & \multicolumn{2}{c}{NCNet \cite{rocco2018neighbourhood}}                 &                      & 2.40                     &                      & 7.61                     &                      & 17.10                     &  & 35.56                 \\
					& \multicolumn{2}{c}{LIFT \cite{yi2016lift}}                 &  & 10.67           &  & 21.19           &  & 31.96        &  & 17.01     \\ 
                   
					& \multicolumn{2}{c}{R2D2 \cite{revaud2019r2d2}+NN \cite{muja2014scalable}}                 &                      & 18.49                     &                      & 35.73                     &                      & 54.80                    &  & 68.66                 \\
					& \multicolumn{2}{c}{DRC-Net \cite{li2020dual}}                 &                      & 20.06                    &                      & 38.16                     &                      & 57.02                   &  & 62.04                \\
                     & \multicolumn{2}{c}{SOSNet \cite{tian2019sosnet}}                 &  & 21.12           &  & 41.80           &  & 56.96     &  & 68.70        \\ 
                    & \multicolumn{2}{c}{MAGSAC \cite{barath2019magsac}}                 &  & 23.87           &  & 42.97           &  & 60.31       &  & 75.71      \\
					& \multicolumn{2}{c}{Patch2pix \cite{zhou2021patch2pix}}                 &                      & 26.25                     &                      & 43.23                    &                      & 59.75                     &  & 72.89                 \\
					
					&\multicolumn{2}{c}{SuperPoint \cite{detone2018SuperPoint}+SuperGlue \cite{sarlin2020SuperGlue}}    &                      & 34.18                    &                      & 50.32                    &                      & 64.16                     &  & 84.90                 \\
					& \multicolumn{2}{c}{LoFTR \cite{sun2021loftr}}                   &                      & $\textbf{37.71}$                    &                      & $\textbf{54.69}$                     &                      & $\textbf{67.00}$                     &  & $\textbf{87.08}$                 \\ 
					\hline
					\multirow{2}{*}{plug-and-play} 	&	\multicolumn{2}{c}{DFM \cite{efe2021dfm}}                    &                      & 15.30                     &                      & 28.57                    &                      & 42.91                     &  & 61.68                 \\
					& \multicolumn{2}{c}{E3CM(ours)}                    &                      & \textcolor{blue}{\textbf{17.63}}                    &                      & \textcolor{blue}{\textbf{31.07}}                     &                      & \textcolor{blue}{\textbf{45.51}}                     &  & \textcolor{blue}{\textbf{76.61}}             
				              \\\hline
			\end{tabular}
                \caption{\textbf{Evaluation on YFCC100M:} The table presents the percentages of correctly estimated poses with pose errors below 5/10/20 degrees. `P' refers to the matching precision in this context.}
		\label{table:yfcc}
	\end{table*}
	
	We also evaluate our proposed method on the MegaDepth dataset \cite{li2018MegaDepth} in terms of pose estimation accuracy. MegaDepth consists of one million internet images from 196 different outdoor scenes and provides ground-truth poses for each image. In addition, it provides sparse reconstructions from COLMAP \cite{schonberger2016structure} and depth maps computed via multi-view stereo approaches.
	
	Following the setup of DISK \cite{tyszkiewicz2020disk} and LoFTR, we specifically focus on the "Sacre Coeur" and "St. Peter's Square" scenes for testing. From these scenes, we select the same 1,500 image pairs as LoFTR for a fair comparison. We estimate the pose using the computed matches by calculating the essential matrix. The pose error is then evaluated by calculating the area under the curve (AUC) of the pose error at thresholds of $5^\circ$, $10^\circ$, and $20^\circ$. The pose error is defined as the maximum angular error in rotation and translation. Although AUC of pose error is influenced by RANSAC, which can discard mismatches and estimate the correct pose, it does not provide a comprehensive evaluation of the matching method. Therefore, we also calculate the matching precision (following SuperGlue \cite{sarlin2020SuperGlue}) as another evaluation metric for correspondence matching.
	
	The evaluation on the YFCC100M dataset follows a similar approach to that on MegaDepth. We select the same image pairs from YFCC100M as used in SuperGlue \cite{sarlin2020SuperGlue} to ensure a fair comparison. We compute the AUC of pose error with thresholds of $5^\circ$, $10^\circ$, and $20^\circ$ and also obtain the matching precision.
	
\begin{figure*}[t!]
		\centering
		\includegraphics[width=0.99\textwidth]{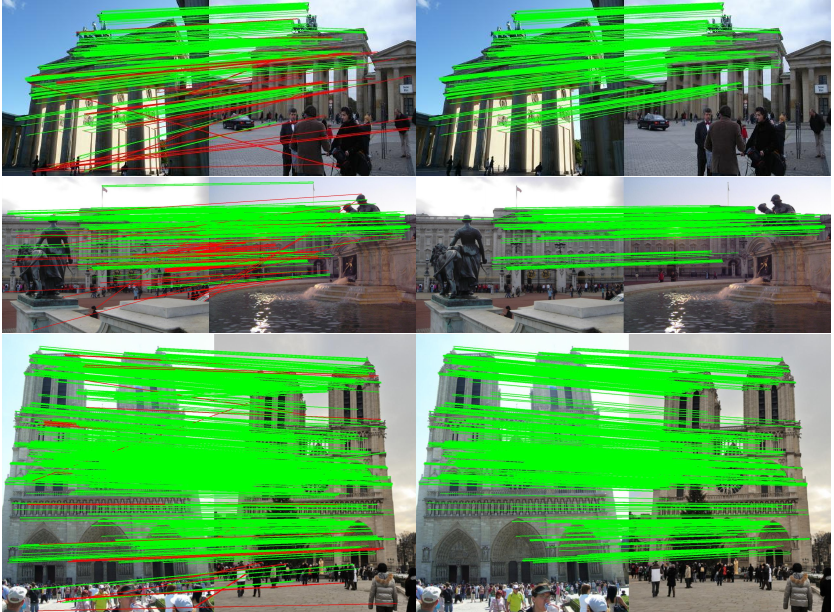}
		\caption{Experimental results on the MegaDepth dataset. Green lines indicate good matches with a Sampson distance below $1 \times 10^{-4}$, while red lines represent bad matches. The images on the left column show the results of direct matching on feature maps without using E3CM, while the images on the right column show the matching results using E3CM.}
		\label{fig:matchingexamples}
	\end{figure*}

	As shown in Tables \ref{table:MegaDepth} and \ref{table:yfcc}, E3CM demonstrates advantages in pose estimation compared to traditional hand-crafted methods and some methods that require training with correspondences. Additionally, our method effectively rejects outliers and achieves high matching precision. As depicted in Fig. \ref{fig:matchingexamples}, the matches in the right column have a much higher matching accuracy rate compared to the matches in the left column, even though both columns have similar pose estimations. While matching accuracy may not be critical in pose estimation, it plays a significant role in other computer vision tasks that rely on correspondence matching, such as stereo rig self-calibration, as discussed in the following subsection.

	\begin{figure*}[t!]
		\centering
		\includegraphics[width=0.98\textwidth]{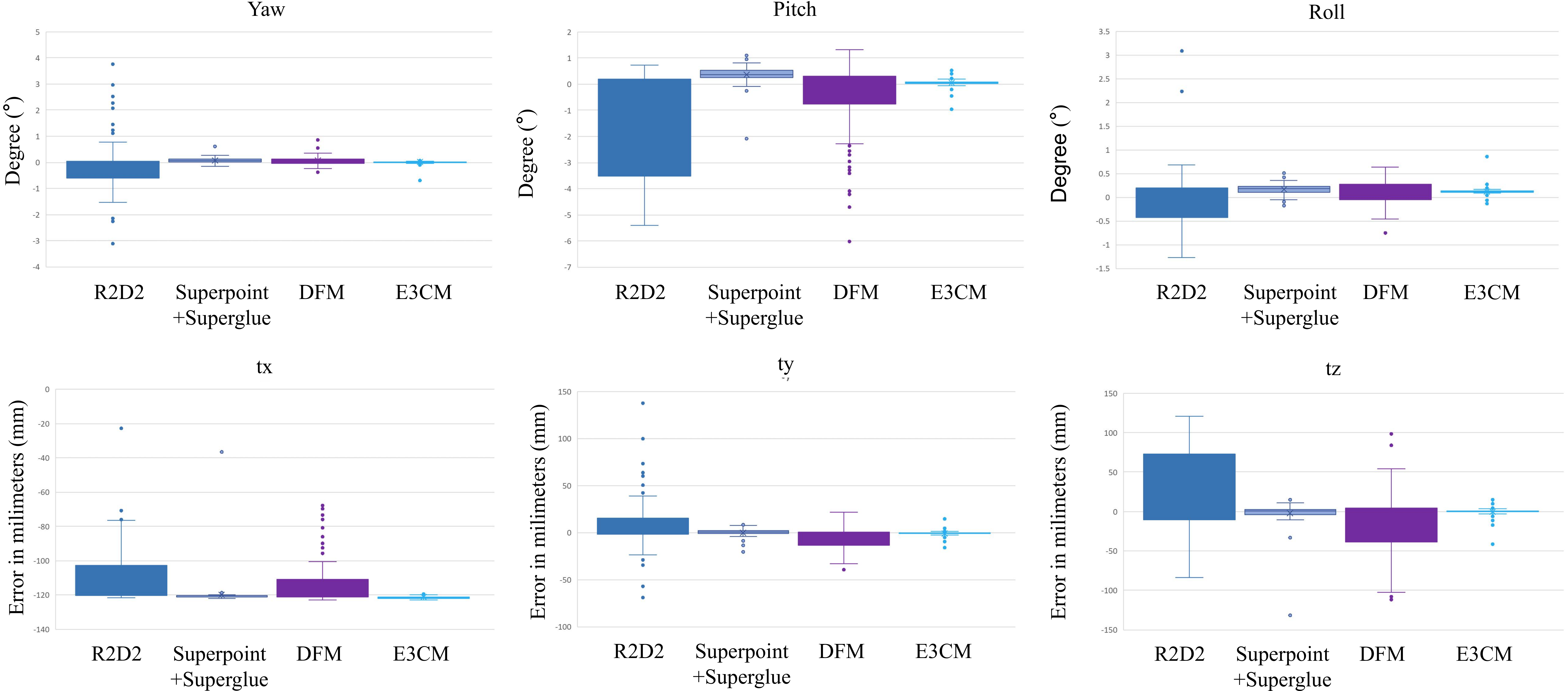}
		\caption{Boxplots of stereo rig self-calibration using different correspondence matching methods. The results obtained using E3CM as the correspondence matching method are the most robust among the four methods. `Yaw', `Pitch', and `Roll' represent the deviation of the rotation angle in three directions. `tx', `ty', and `tz' represent the translation error in three directions.}
		\label{fig:boxplot}
	\end{figure*}

        \begin{figure}[t!]
		\centering
		\includegraphics[width=0.5\textwidth]{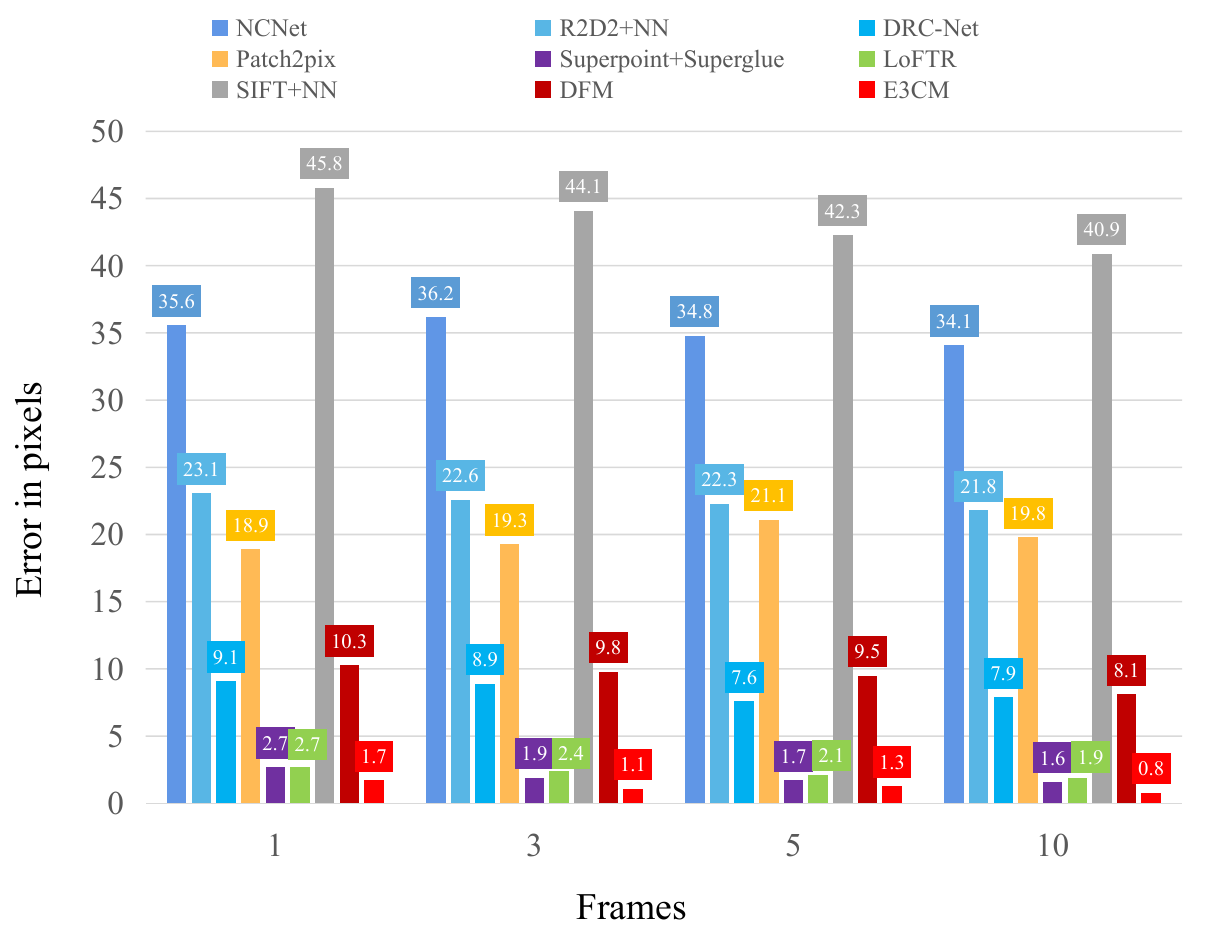}
		\caption{Checkerboard's average reprojection errors, obtained using different correspondence matching methods. }
		\label{fig:reprojectionerror}
	\end{figure}

 \begin{figure*}[t!]
		\centering
		\includegraphics[width=0.99\textwidth]{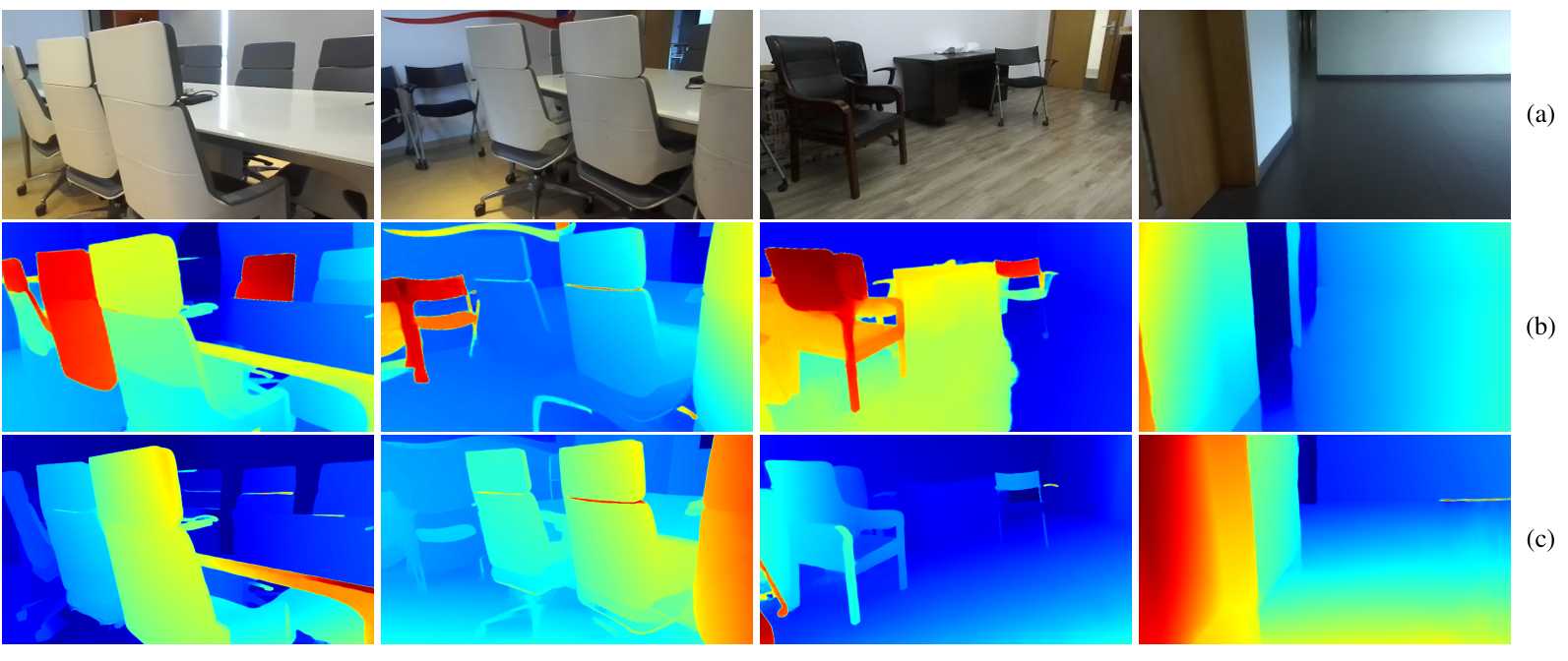}
		\caption{Stereo rig self-calibration results: (a) reference RGB images; (b) disparity images for the unrectified stereo image pairs; (c) disparity images for the rectified stereo image pairs which are obtained via stereo rig self-calibration. The disparity maps are computed using RAFT-Stereo \cite{lipson2021raft}. }
		\label{fig:disparity}
	\end{figure*}

	\subsection{Stereo rig self-calibration}
	
	Among the various computer stereo vision tasks that require correspondence matching, stereo rig self-calibration is of particular importance. Stereo rig self-calibration heavily relies on highly accurate correspondence matching. Even a small portion of bad matches can lead to significant deviations in the calibration results. The stereo rig self-calibration pipeline \cite{ling2016high} typically involves correspondence matching in the first stage and the optimization of extrinsics based on the matches and epipolar constraints. Hence, we can evaluate the accuracy of front-end correspondence matching methods based on the results of the back-end stereo rig self-calibration.
	
	Following \cite{ling2016high}, we replace the corner detector \cite{shi1994good} and BRIEF \cite{calonder2010brief} with other keypoint extraction and matching methods based on deep neural networks. Additionally, we perform stereo rig self-calibration in several scenes. To enhance calibration stability, we input all matches within ten frames into the optimizer in our experiments. The calibration results are presented as boxplots, a common evaluation approach in the calibration domain, as shown in Fig. \ref{fig:boxplot}. Our method exhibits fewer outliers and smaller ranges in the calibration results compared to other methods, indicating that our plug-and-play method is more accurate and robust, even outperforming some fully supervised approaches.
	
	Stereo rig calibration is a critical step in terms of achieving accurate 3D measurements \cite{fan2018road,fan2022rethinking}. Therefore, we further compute the reprojection error of checkerboard pattern corners with respect to different calibration results obtained using 1, 3, 5, and 10 frames, respectively. We compare the performance of several state-of-the-art methods, including our proposed E3CM and SuperPoint+SuperGlue. The results shown in Fig. \ref{fig:reprojectionerror} suggest that E3CM outperforms SuperPoint+SuperGlue in all calibration set-ups, achieving a significantly lower reprojection error.
	
	To provide a more intuitive evaluation, we generate dense disparity maps using RAFT-Stereo \cite{lipson2021raft}, as shown in Fig. \ref{fig:disparity}. We generate disparity maps for both rectified and unrectified stereo image pairs. It is evident from Fig. \ref{fig:disparity} that the unrectified disparity maps yield erroneous depth estimations, while the well-rectified disparity maps obtained using our proposed correspondence matching method accurately reflect the depth relationships between different objects in the scenes.
 
 	\section{Discussion}
  \label{sec.discussion}
     
    Our method is well-suited for stereo vision systems. However, for monocular vision systems that require correspondence matching between consecutive frames, the accuracy of our method may be significantly affected if there are dynamic objects in the scene. This is because our method relies on the assumption of a stationary background with features, such as a building, while a moving object is present. In such cases, our method may estimate incorrect poses during the epipolar-constrained cascade refinement. Specifically, in monocular cameras, if points from both stationary and dynamic objects are selected simultaneously during pose estimation, the relative pose estimation between the former and latter frames of the monocular camera will be erroneous. This discrepancy arises because monocular camera pose estimation computes the camera-to-scene relationship, and when the scene contains moving objects, it introduces changes that invalidate the pose estimation process. In contrast, stereo vision systems have relatively static cameras with respect to each other. Therefore, this problem does not arise in stereo vision systems since the pose estimation is computed from camera to camera and is independent of the scene. Although our method is somewhat limited to stereo vision systems, stereo vision remains a critical aspect of computer vision, and our method can still find broad applications in this context.
	
	The epipolar-constrained cascade refinement method is effective for pose estimation and outlier rejection. When matching directly in the source image, the limited feature information per pixel often leads to numerous mismatches with similar features but different spatial locations. Consequently, using these matches directly for pose estimation can result in incorrect pose estimations. Moreover, removing outliers based on these estimated poses would propagate incorrect matches further. In contrast, matches in the deep feature map, which are convoluted from patches in the source image, contain more semantic information \cite{fan2020sne}. As a result, the probability of mismatches is significantly reduced compared to direct matching in the source image. Although the pose estimation based on these points, which correspond to patches in the source image, may not be highly accurate, it is generally reasonable and relatively correct. As the number of layers decreases, the coordinates of the points corresponding to the patches in the source image become more accurate. Consequently, the pose estimation based on these points becomes more precise, and the matches obtained by removing outliers based on the estimated pose also improve in accuracy. In essence, our proposed method ensures that the initial pose estimation is reasonable and progressively becomes more accurate. Therefore, outlier rejection based on the initial pose estimation is effective and accurate. Additionally, estimating a pose during the epipolar-constrained cascade refinement requires a minimum of eight pairs of matched points. However, in our experiments, we rarely encountered cases with fewer than eight pairs.
	
	We compute the fundamental matrix by mapping the points in the feature maps to the corresponding pixels in the source image. This process does not require camera intrinsics. However, if camera intrinsics are available, we can directly calculate the essential matrix in the feature maps without the need for mapping the points to the source image pixels. In our experiments, we observe a slight improvement in accuracy using this approach, although the improvement is not significant. The reason is that mapping the points in the feature maps to the pixels in the source image and then computing the fundamental matrix is equivalent to pose estimation based solely on the pixels, which ignores the information from other pixels in the patch. On the other hand, computing the essential matrix directly in the feature maps is equivalent to pose estimation based on the patches in the source image, which includes more information and can result in improved accuracy.
 
	\section{Conclusion}
	\label{sec.conclusion}
	
	In this paper, we introduced a new plug-and-play correspondence matching method that leverages an epipolar-constrained cascade refinement strategy. Our approach is compatible with various pre-trained backbone networks, allowing for flexibility in choosing the most suitable backbone for the task. We argue that pre-trained networks from other vision tasks can be effectively utilized in the correspondence matching task, often achieving comparable or even superior performance compared to fully supervised methods specifically designed for correspondence matching. By leveraging the representation learning capabilities of pre-trained networks, we can leverage the rich knowledge acquired from large-scale datasets, enabling efficient and accurate correspondence matching. Through our experiments and evaluations, we have demonstrated the effectiveness of our approach and its ability to leverage pre-trained networks for correspondence matching tasks. We believe that our method opens up new possibilities for utilizing pre-trained networks in a broader range of vision tasks, offering improved performance and efficiency.
	
%	In this paper, we presented a novel plug-and-play correspondence matching method based on an epipolar-constraint cascade refinement strategy. Our approach works for a variety of pre-trained backbone networks. We believe that for the correspondence matching task, pre-trained networks used in other vision tasks can be fully utilized to achieve the same or even better performance than some fully supervised correspondence matching methods which are specifically designed for this task. 

%	\clearpage
	\bibliographystyle{model1-num-names}

	\end{document}